\documentclass[sigconf,nonacm]{acmart}
\setcopyright{none}
\renewcommand\footnotetextcopyrightpermission[1]{}
\acmConference[SAC '27]{The 42nd ACM/SIGAPP Symposium on Applied Computing}{April 5--9, 2027}{Gwangju, South Korea}
\acmYear{2027}
\copyrightyear{2027}

\usepackage{booktabs}
\usepackage{flafter}
\usepackage{tabularx}
\usepackage{tikz}
\usetikzlibrary{arrows.meta}
\usepackage{listings}
\usepackage{xcolor}
\begin{document}

\title{RoboFACTS: A Declarative Experiment Design Framework for Multi-Robot Trials in Simulation and on Hardware}

\author{Alejandro Lorite-Mora}
\orcid{0000-0002-8704-1248}
\affiliation{\institution{IT University of Copenhagen}\city{Copenhagen}\country{Denmark}}
\additionalaffiliation{\institution{Novo Nordisk A/S}\city{Kalundborg}\country{Denmark}}
\email{lori@itu.dk}

\begin{abstract}
Results in multi-robot research come from trials, in simulation and with physical robots. The conditions of each trial are often tracked by hand, so a trial that did not run as declared can still count as evidence and change the reported result without anyone noticing. We present RoboFACTS (Factors, A Contract, and checked Trials in Simulation and on hardware), a declarative experiment design framework. The researcher declares the factors of an experiment, their levels, and the fields that every trial must record. The framework plans the trials that are still missing, checks the setup before each trial and the recording after it, and counts only the trials that pass, in simulation and on hardware. We evaluated it with ROS 2 in four studies, in our laboratory with a quadruped robot and small drones and in simulation. In the laboratory, the checks rejected trials whose software had exited normally, and counting them would have raised the measured error of one condition by 7\,\%. With faults injected on real robots and in simulated teams of up to 20 robots, the checks refused every faulty attempt, and in the teams the record named the faulty robot in almost every case. Faults that no working check covered were counted and changed the result. The share of laboratory recordings with a complete record rose from 6.3\,\% to 86.2\,\%, in part because no contract existed before. Large language models (LLMs) matched the requested design in every declaration but not in every free-form script, and an LLM coding agent declared and ran a new experiment from the documentation alone. A counted trial can be wrong only in a factor that no working check covers, and every failure we observe becomes a new check. The framework is open source, and its core fits any experiment that software runs, because it needs only a command that starts the experiment, a recorder, and a script for each check.
\end{abstract}

\ccsdesc[500]{Computer systems organization~Robotics}
\ccsdesc[300]{Computing methodologies~Multi-agent systems}

\keywords{multi-robot systems, multi-agent systems, design of experiments, reproducibility, ROS 2}

\maketitle

\section{Introduction}
\label{sec:intro}

\begin{figure}[t]
    \centering
    \resizebox{\columnwidth}{!}{\begin{tikzpicture}[
    font=\sffamily\scriptsize,
    lane/.style={draw=black!20, fill=black!3},
    box/.style={draw, rounded corners=1.5pt, align=center, inner sep=2pt, minimum height=0.62cm, fill=white},
    decl/.style={box, fill=blue!8},
    check/.style={box, fill=orange!12},
    rec/.style={box, fill=green!10},
    stopb/.style={box, fill=red!8},
    arr/.style={-{Stealth[length=1.6mm]}, thick},
    lbl/.style={font=\sffamily\tiny, text=black!65},
    lane label/.style={font=\sffamily\tiny\bfseries, text=black!55, align=right},
]
\def\W{9.5}
\foreach \y in {0, -1.2, -2.4} { \fill[lane] (0,\y+0.55) rectangle (\W,\y-0.55); }
\node[lane label, anchor=north east] at (\W-0.03,0.55) {Operator};
\node[lane label, anchor=north east] at (\W-0.03,-0.65) {Control computer};
\node[lane label, anchor=south east] at (\W-0.03,-2.93) {Trial computer\\(lab PC or robot)};
\foreach \x/\t in {1.1/Declare and plan, 2.7/Gate, 4.35/Run, 5.95/Verify, 8.15/Count} {
    \node[font=\sffamily\tiny\bfseries, anchor=base] at (\x,-3.3) {\t};
}
\node[decl, text width=1.55cm] (decl) at (1.1,0) {declare factors\\and campaign};
\node[decl, text width=1.55cm] (camp) at (1.1,-1.2) {plan the trials\\still missing};
\node[check, minimum width=1.2cm, minimum height=3.3cm] (gate) at (2.7,-1.2) {};
\node[align=center] at (2.7,0) {operator\\confirms};
\node[align=center, lbl] at (2.7,-1.2) {result stored\\with the trial};
\node[align=center] at (2.7,-2.4) {automatic\\checks};
\node[stopb, text width=1.25cm] (stop) at (4.35,0) {operator\\stop};
\node[rec, text width=1.25cm] (run) at (4.35,-2.4) {trial runs,\\data recorded};
\node[check, text width=1.2cm] (ver) at (5.95,-2.4) {check the\\recording};
\node[stopb, text width=1.2cm] (quar) at (5.95,0) {excluded,\\with reason};
\node[rec, text width=1.45cm] (idx) at (7.45,-1.3) {coverage index};
\node[decl, text width=0.85cm] (ana) at (8.95,-1.3) {analyses};
\draw[arr] (decl) -- (camp);
\draw[arr] (camp.east) -- (gate.west |- camp.east);
\draw[arr] (gate.east |- run.west) -- node[lbl, above] {GO} (run.west);
\draw[arr, dashed] (stop.south) -- node[lbl, right, pos=0.5] {abort} (run.north);
\draw[arr, dashed] (stop.east) -- (quar.west);
\draw[arr] (run) -- (ver);
\draw[arr] (ver.north) -- node[lbl, right, pos=0.55] {fail} (quar.south);
\draw[arr] (ver.east) -- (7.1,-2.4) -- node[lbl, right, pos=0.45] {pass} (7.1,0 |- idx.south);
\draw[arr] (idx) -- (ana);
\draw[arr] (idx.north) -- (7.45,0.78) -- node[lbl, above, pos=0.5] {next campaign: the conditions still missing} (1.1,0.78) -- (decl.north);
\end{tikzpicture}}
    \Description{A flow diagram in three rows: the operator, the control computer, and the trial computer. The operator declares the factors and the campaign, and the control computer plans the trials still missing. Before each trial, automatic checks run and the operator confirms, and the result is stored with the trial. The trial computer runs the trial and records the data, and the operator can stop it. Then the recording is checked: a trial that fails is excluded with a reason, and a trial that passes enters the coverage index, which feeds the analyses and the next campaign.}
    \caption{The life cycle of one trial. Rows are the operator, the control computer, and the computer that runs the trial. The planner selects a missing trial, the gate checks the setup, the verify steps check the recording, and only a trial that passes both enters the coverage index, which guides the next campaign.}
    \label{fig:loop}
\end{figure}
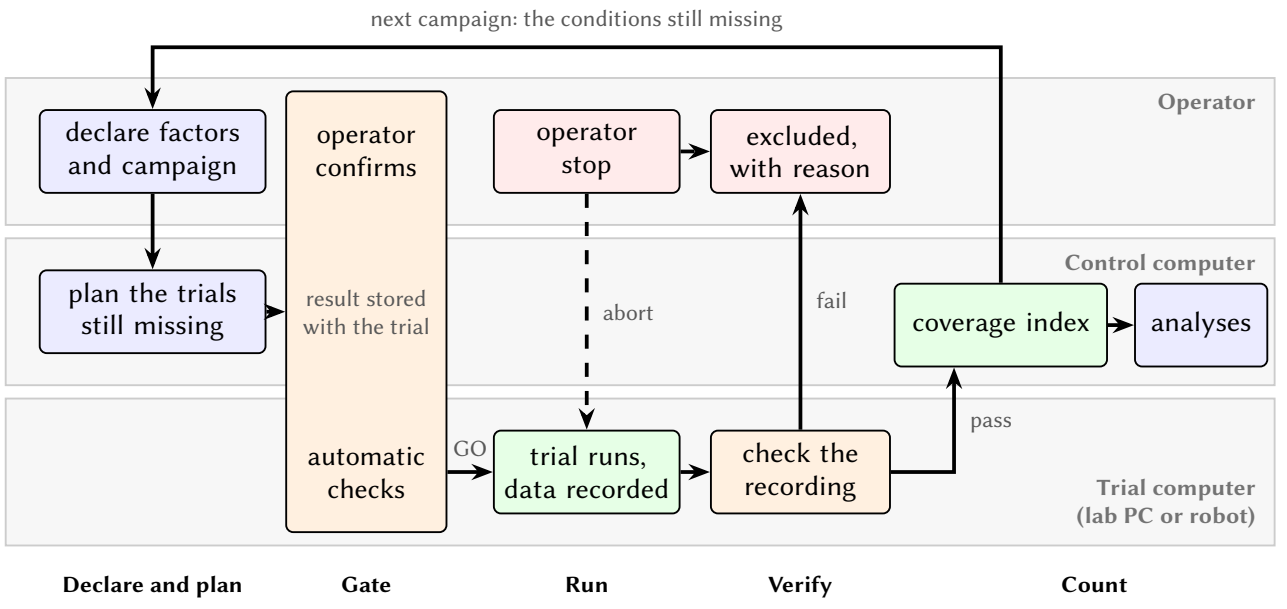

Many results in multi-robot research come from trials, in simulation or with physical robots. The community has described in detail what a trial should report: the data sets, code identifiers, and hardware of a replicable paper \cite{bonsignorio2015toward}, the conditions, number of runs, and success criteria of an evaluation of a learned robot policy \cite{kressgazit2024robot}, and guidelines for runtime verification and field-based testing of ROS-based systems \cite{caldas2024runtime}. None of these works captures and checks the conditions of each trial at the moment the trial is recorded, when this information is kept or lost.

A field report by Ortega et al. shows what happens when this information is managed by hand \cite{ortega2022testing}. During a ten-month test campaign with a service robot, the team filled in its test-report schema only occasionally, because the manual effort was too high. Only 7\,\% of the recorded bag files were listed in a test report, and the authors recommend that teams ``automate the data collection from field tests, including data provenance and traceability of requirements and test cases.'' We met the same problem in our own laboratory, where a quadruped robot and up to two nano unmanned aerial vehicles (nano-UAVs) run joint localization experiments with three computers and one operator. From March to July 2026 we recorded there without a framework, and the failures we observed motivated this work.

For simulation, such tools now exist. RoboVAST declares a validation campaign in one file and runs thousands of scenario configurations on a cluster \cite{pasch2026robovast}, and Ortega et al. record the provenance of every run \cite{ortega2026replicable}. Both target simulation and a single robot. Trials with a team of physical robots, such as ours, add three problems. First, a human operator prepares the robots and confirms conditions that no script can check. Second, each trial involves several agents, robots and computers, and one faulty agent can spoil it, so the record must name which agent did what. Third, a process can exit without error even though the robot never performed its task, so the recording itself must be checked. In each case, a trial can count as evidence although it did not run as declared, a \emph{silent error}. Results in simulation also do not always transfer to a real robot \cite{kadian2020sim2real}, so the record must show which conditions have evidence only from simulation.

This paper presents RoboFACTS, a declarative experiment design framework that records and checks this information during each trial, in simulation and on hardware (Figure~\ref{fig:loop}). The name stands for Factors, A Contract, and checked Trials in Simulation and on hardware. A team declares an experiment once, validates it in simulation, and spends its hardware time on the conditions that only hardware can answer, while the record keeps the two kinds of evidence apart. We use the terms of the design of experiments. A \emph{factor} is a parameter that the experiment controls, such as the number of drones or the lighting, and its \emph{levels} are the values it can take. The researcher declares the factors and a \emph{metadata contract}, the list of fields that every trial must record. A \emph{campaign} selects which factors to vary and how many times to repeat each trial, and new campaigns extend the experiment as the work progresses. The framework plans the remaining trials, runs a \emph{gate} of machine checks and operator confirmations before each trial, runs \emph{verify steps} on the recording after each trial, and counts only the trials that pass, the \emph{usable} trials, in a \emph{coverage index}. A counted trial can then carry a silent error only in a factor that no check covers (Equation~\eqref{eq:silent}). The framework makes these factors visible, and every failure we observe becomes a new check.

The paper makes three contributions. First, a framework that a researcher uses by declaring an experiment, as a general package and a package that connects it to ROS 2. Second, three features that we did not find together in prior work: a metadata contract that is checked against the factors before any trial runs, a coverage index that lists the conditions without a usable trial, and a gate that stores machine checks and operator confirmations with the trial. Third, an evaluation in four studies. Every check that fired under injected faults stopped its fault, with one faulty agent among up to 20 robots the record named it, and in a real laboratory complete records rose from 6.3\,\% to 86.2\,\% of recordings. Declarations are short files that validators check, so they are also a safe target for large language models (LLMs) that design experiments, which a fourth study tests. RoboFACTS, three example experiments, and the documentation are available at \url{https://github.com/Lorite/robofacts}.

Section~\ref{sec:related} compares the framework with related work. Section~\ref{sec:loop} describes how a researcher uses it, its formal model, and its mechanisms. Section~\ref{sec:eval} reports the four studies, and Section~\ref{sec:discussion} discusses them and the limitations.

\section{Related Work}
\label{sec:related}

Table~\ref{tab:features} compares six works with ours on ten features. The first five columns are the features of Section~\ref{sec:intro}: a machine-readable factor design, the contract, the coverage index, the gate, and the verification of each recording, which removes a failing trial from the count. A feature is present (Y) only in the full sense of our contributions. A feature is partial (P) when it exists only in prose, only after the campaign, or without some checks, for example, a gate with machine checks only. The other columns mark execution on several computers, trials with more than one physical robot, versioned analyses, a provenance graph, and a user study. None of the other works runs hardware trials with an operator, so we compare features.

\begin{table}[htbp]
    \caption{Feature coverage of the related work, verified in the full texts. Y: present, P: partial, N: absent.}
    \label{tab:features}
    \footnotesize
    \setlength{\tabcolsep}{2.2pt}
    \begin{tabular}{@{}lcccccccccc@{}}
        \toprule
        Work & Fact. & Contr. & Cov. & Gate & Verif. & Hosts & Rob. & Anal. & Prov. & Study \\
        \midrule
        Field report \cite{ortega2022testing} & P & P & P & N & N & N & N & N & N & N \\
        CITK \cite{lier2016towards} & N & P & N & N & P & P & N & P & N & Y \\
        Eurobench \cite{remazeilles2022making} & P & P & N & N & N & N & N & Y & N & N \\
        Autolab \cite{tani2020integrated} & N & P & N & P & P & P & P & Y & N & N \\
        RoboVAST \cite{pasch2026robovast} & Y & P & P & N & P & Y & N & P & N & N \\
        Ortega et al. \cite{ortega2026replicable} & Y & P & N & N & N & Y & N & P & Y & N \\
        This paper & Y & Y & Y & Y & Y & P & Y & Y & N & N \\
        \bottomrule
    \end{tabular}
\end{table}

\paragraph{Reproducibility toolchains for hardware} The Cognitive Interaction Toolkit (CITK) \cite{lier2016towards} combines a versioned system description, deployment through continuous integration, and an executable experiment description. Its unit is a single fixed experiment on a single computer, and its three study systems were simulated or replayed. Eurobench \cite{remazeilles2022making} defines a file format for each run and packages 17 performance-indicator algorithms as Docker images, but the format applies to data uploaded after the experiment. The Duckietown Autolab \cite{tani2020integrated} standardizes the laboratory itself, with identical robots and a hardware compliance test before each run, and its only variable is the submitted agent.

\paragraph{Campaign frameworks in simulation} RoboVAST \cite{pasch2026robovast} defines coverage as the share of the valid scenario space that a campaign has covered, and Ortega et al. \cite{ortega2026replicable} store the provenance of each run as a graph in PROV-O, the W3C standard for provenance. Neither paper involves an operator, declares a per-trial contract, or quarantines failed trials. We adopt their definition of coverage and add the parts they identify as missing for physical trials.

\paragraph{Experiment managers and ROS 2 tools} Machine-learning experiment managers such as Hydra, Sacred, and MLflow track the configurations and results of runs. These tools do not model the computer that runs a trial, the operator, or checks on the recorded data.

\paragraph{LLMs that write robot programs and experiments} Code as Policies generates Python policy code from instructions \cite{liang2023code}, and Vemprala et al. prompt ChatGPT with a curated library of high-level functions and keep a user ``on the loop'' who checks the generated code before it runs on the robot \cite{vemprala2024chatgpt}. Coscientist \cite{boiko2023autonomous} and the AI Scientist \cite{lu2026towards} let an LLM design an experiment, write and run its code, and repair errors or interpret results only after execution. CLAIRify lets an LLM write programs in a structured language for chemistry robots and returns a verifier's errors for another attempt \cite{yoshikawa2023large}, and Kambhampati et al. argue that LLMs should plan only together with external verifiers \cite{kambhampati2024llm}. Yang et al. propose a declarative stack for self-driving laboratory experiments and demonstrate it on one experiment in a battery laboratory, but no language model writes the declarations and the benefits are not measured \cite{yang2026experiment}. Our framework adds operator confirmations, recording checks, and a coverage index, and Section~\ref{sec:rq3} measures how the declarations change LLM designs.

\paragraph{Coverage of conditions in simulation and on hardware} Rigorous measures of test adequacy and public tools are rare in robot testing \cite{araujo2023testing}. Situation coverage measures how much of the environmental situation space a test campaign has exercised \cite{alexander2015situation,majzik2019towards}. Navigation results in simulation and on a real robot correlated at only 0.18 under the settings of a 2019 challenge, and at 0.84 after tuning the simulator \cite{kadian2020sim2real}, and test results did not always transfer between a small car and its digital twin \cite{stocco2023mind}. Scenario-based testing moves selected scenarios from simulation to a test track \cite{fremont2020formal}, or rebuilds a track scenario in simulation to measure the gap between the two \cite{reway2020test}. Fremont et al. compare the declared and the realized scenario parameters, but only after the campaign and without excluding trials. To our knowledge, none of these works checks the declared condition of a hardware trial before counting it, or reports per level whether the evidence comes from hardware or only from simulation, which Equations~\eqref{eq:silent} and~\eqref{eq:hgap} make explicit.

\section{The Framework}
\label{sec:loop}

\subsection{Workflow: Declare, Plan, Run, and Count}
\label{sec:usage}

A researcher uses the framework in four steps. First, they declare the experiment in a few short YAML and JSON files (Listing~\ref{lst:declaration}): the factors and their levels, the metadata contract, the gate, the runner configuration, and the named analyses. Second, they write a campaign, a short file that selects the factors to sweep, fixes the others, and sets the number of repetitions. Before any trial runs, a linter checks the declarations and the campaign, for example, that every swept factor is recorded. Third, they start the campaign from the command line or a web interface. The planner lists the trials the coverage index still lacks, and for each one, the runner starts the software on the right computer, runs the gate, records the trial, and runs the verify steps (Figure~\ref{fig:loop}). Fourth, they read the coverage index, which shows how many usable trials exist per condition, and run the named analyses, which select their trials from the index. A trial that failed a check stays visible with its reason, but it never counts.

\begin{lstlisting}[float=htbp,caption={The example experiment (no robot) from declaration to coverage: excerpts of its factor file, contract, gate, and campaign, and two of the four cells of its coverage index, shortened and reformatted.},label={lst:declaration}]
# config/factors.yaml
factors:
  execution_host:
    levels: [laptop, lab_pc, rig]
    metadata_key: execution_host
    maps_to: {kind: host}
  chatter_rate:
    levels: [slow, fast]
    metadata_key: chatter_rate
    maps_to: {kind: internal}

# config/trial_metadata_schema.yaml (the contract)
fields:
  chatter_rate: {required: true, factor: true, enum: [slow, fast]}

# config/preflight_checklist.json (the gate)
{"text": "The bag directory is writable.", "check": "bag_dir_writable"},
"Nothing physical to check: this demo has no robot."

# campaigns/rates.yaml
repetitions: 2
sweep:
  chatter_rate: [slow, fast]

# the coverage index after the campaign
chatter_rate | execution_host | usable | unverified | excluded
fast         | lab_pc         |      4 |          0 |        0
slow         | laptop         |      2 |          0 |        0
\end{lstlisting}

\subsection{Formal Model of Checks and Coverage}
\label{sec:model}

This section defines the objects of the framework. An \emph{experiment} is a physical setup together with its software, for example, the quadruped observing the drone. It declares a set of factors $F$. Each factor $f \in F$ has a finite set of levels $L_f$, a default $d_f \in L_f$, a metadata key $\kappa(f)$, and a realization: the computer that runs the trial, a launch argument, a recording profile, an internal value, or an operator step. The condition space is $\Omega = \prod_{f \in F} L_f$, and a \emph{cell} is one condition $\omega \in \Omega$. A campaign $c$ sweeps the factors $S_c \subseteq F$ over levels $L^c_f \subseteq L_f$, fixes every other factor at a value $x_f$, and asks for $r_c(\omega)$ repetitions of each cell of the full product
\begin{equation}
\Omega_c = \prod_{f \in S_c} L^c_f \ \times \prod_{f \in F \setminus S_c} \{x_f\}.
\label{eq:cells}
\end{equation}
A \emph{trial} $t$ runs for a declared condition $\delta(t) \in \Omega_c$ and produces a recording $b(t)$ and a metadata record $m(t)$. Its realized condition $\rho(t)$ is the condition that actually held during the trial, which the framework observes only through checks. A trial has a silent error when it counts as coverage although the condition that actually held differs from the declared one, $\rho(t) \neq \delta(t)$, for example, a hover recorded in still air that ran in the windy world.

\paragraph{Contract and linter} The contract declares the keys $K$ that every record must contain and the coverage axes $X \subseteq K$. The rule that the planner, the campaign checker, and the runner enforce is the invariant
\begin{equation}
\forall f \in S_c: \ \kappa(f) \in X,
\label{eq:lint}
\end{equation}
and a campaign that violates it is refused before any trial runs. Under this invariant, the cell of a trial can be recovered from its record alone, as the projection of $m(t)$ onto the coverage axes, so it never depends on a notebook or a file name.

\paragraph{Gate and verify steps} The gate is a set of checks $G$ evaluated immediately before a trial. A machine check is a predicate $g(s, \delta(t))$ over the observed state $s$ of the computers and robots, and it may depend on the declared level (for example, the expected simulator world). The trial starts only if every binding check (a check that blocks) holds. Verify steps $V$ are predicates $v(b(t))$ on the recording, evaluated after the trial, and a trial that fails one is aborted.

\paragraph{Classes and coverage} Every trial receives one class,
\begin{equation}
\mathrm{cls}(t) = \begin{cases} \text{excluded} & \text{aborted, quarantined, or no data,} \\ \text{unverified} & \text{else, if } m(t) \text{ does not satisfy the contract,} \\ \text{usable} & \text{otherwise.} \end{cases}
\label{eq:class}
\end{equation}
The coverage of a cell counts its usable trials, $n_c(\omega) = |\{t : \delta(t) = \omega, \ \mathrm{cls}(t) = \text{usable}\}|$, and the planner returns the cells that still lack trials, $M_c = \{\omega \in \Omega_c : n_c(\omega) < r_c(\omega)\}$.

\paragraph{Realized conditions} A check is \emph{sound} for a factor $f$ if it passes only when $\rho_f(t) = \delta_f(t)$. Let $F_G \subseteq F$ be the factors that have such a check, in the gate or in a verify step. If a factor in $F_G$ differs, its check fails and the trial is blocked or aborted, so a usable trial matches its declaration in every checked factor:
\begin{equation}
\mathrm{cls}(t) = \text{usable} \ \Rightarrow \ \forall f \in F_G : \rho_f(t) = \delta_f(t).
\label{eq:silent}
\end{equation}
A silent error can therefore lie only in $F \setminus F_G$, and the framework makes this set visible: a factor declared as realized by an operator step, such as the lighting of the room, must declare a check or a reason for having none, or the linter refuses it.

\paragraph{Simulation and hardware} Each trial runs in a domain $D(t) \in \{\text{sim}, \text{hw}\}$, which the runner records and which older trials derive from their experiment's declaration. A trial counts as evidence for a level $\ell$ only if it is usable and the realization check of $f$ at its gate passed, so a usable trial whose condition was never checked is reported but not counted. With $n^{D}_f(\ell)$ the number of such trials per level and domain, the framework reports the levels that have evidence only from simulation,
\begin{equation}
H_f = \{\ell \in L_f : n^{\text{sim}}_f(\ell) > 0, \ n^{\text{hw}}_f(\ell) = 0\}.
\label{eq:hgap}
\end{equation}
A condition that is easy to vary in simulation but hard to reproduce on hardware, such as lighting, thus appears in $H_f$ until a hardware trial with a passing realization check covers it.

\subsection{Mechanisms and the Failures Behind Them}
\label{sec:mechanisms}

The mechanisms below follow the life cycle of Figure~\ref{fig:loop}. Each one began with a failure in our own trials in which the software reported a normal trial that was not valid, and each failure produced a rule for every later experiment (Table~\ref{tab:incidents}).

\begin{table}[htbp]
    \caption{Failures observed in our laboratory's trials and the mechanism each one produced.}
    \label{tab:incidents}
    \footnotesize
    \begin{tabularx}{\columnwidth}{@{}XX@{}}
        \toprule
        Observed failure & Mechanism added \\
        \midrule
        An emergency stop was still latched, and the trial recorded 48\,s of a grounded drone. & The gate reads the drone's supervisor state before every trial. \\
        A flight script exited with code 0, but no flight took place. & Every experiment must declare verify steps on the recording, or state why it has none. \\
        A verify step judged the whole recording and marked a clean 68\,s hover as aborted. & Verify steps find the airborne window from the flight-state messages in the recording. \\
        An internal factor flew the drone at 0.5\,m in a campaign that declared 1.0\,m. & The linter refuses an internal factor that no command uses. \\
        A campaign swept a level whose equipment was not installed. & A level can be declared blocked, and a campaign that uses it is refused at planning time. \\
        An operator stop killed the observer node, and the trial continued without it. & An operator stop aborts the trial, and the trial never resumes. \\
        \bottomrule
    \end{tabularx}
\end{table}

\paragraph{Declared files} An experiment is declared in five short files (Listing~\ref{lst:declaration}): the factors with their levels, defaults, and realizations, the metadata contract with its coverage axes, the gate checklist, the runner file, which also names the verify steps and how to reach each computer, and the analysis file.

\paragraph{Campaigns} Besides the sweep, a campaign declares the conditions that its trials must meet: the steps the operator reads before a trial, the steps shown while it runs, its own verify steps, and its own gate when its preconditions differ from the experiment's. Such a gate lists what must hold for that campaign. If the campaign reused the experiment's gate, some items would fail in every trial, and an operator who overrides expected failures will in time override a real one.

\paragraph{The linter} The most important rule connects the factor file with the contract: every swept factor must name the metadata field in which it is recorded, and that field must be a coverage axis (the \texttt{metadata\_key} line and the contract entry of Listing~\ref{lst:declaration}), because the coverage of an unrecorded factor can never be measured. The planner, the campaign checker, and the runner refuse a campaign that breaks this rule. The configuration linter checks every experiment against sixteen such rules, for example, a launch argument that the launch file does not declare or an experiment without verify steps.

\paragraph{The gate} The gate runs before every trial and never reuses a previous result. Each machine check is a script, named in its item of the checklist, and an item without one is an operator confirmation (Listing~\ref{lst:declaration}). For example, the gate checks that no processes remain from a previous run, that the recording directory is writable, that the drone's battery and supervisor state are correct, and that motion capture tracks every rigid body. The operator confirms the remaining items of the same checklist in the terminal or browser.

\paragraph{The gate record} The gate writes its result into the trial's metadata (Listing~\ref{lst:gate}), including the failed items and any overrides with their reasons, so the operator is part of the record. The gate also appends every verdict, including rejections that leave no trial behind, to a persistent gate log, and records every \emph{attempt}, a started trial whether the gate refused it or not, in a ledger.

\begin{lstlisting}[float=htbp,caption={The gate record in the metadata file of one trial of the fault study (Section~\ref{sec:rq1}), an excerpt. The non-blocking gate, named \texttt{off} in the files, ran its 8 checks and stored its NO-GO, and a later layer aborted the trial. An operator's override stores its reason in the same field.},label={lst:gate}]
# trial 20261002T001156_fault-dds_no_discovery_gate-off_rep04
"preflight": {
  "verdict": "NO-GO", "n_checks": 8, "duration_s": 7.24,
  "checked_at": "2026-10-02T00:12:42+00:00",
  "failed_checks": ["topic_rate: /fmu/out/vehicle_local_position_v1 is not advertised at all"],
  "overridden": true,
  "override_reason": "shadow: gate=off (the gate runs and records, a NO-GO does not block)"
},
"aborted": true,
"abort_reason": "action returned rc=2 ..."
\end{lstlisting}

\paragraph{Operator stop} An operator stop writes a latch file, and the trial script then aborts, attempts to land the drone, and refuses to start again while a latch younger than 120\,s exists.

\paragraph{Verify steps} Verify steps are executable checks on the recording that test the effect of a trial, whatever its cause, for example, whether the drone left the ground and whether the vision estimate fed the control loop while it was in the air. The runner applies the same steps to every trial, and a campaign can add its own. When a step fails, the runner marks the trial as aborted, using the step's output as the reason, and excludes it from the index. The operator moves a failed trial into a quarantine folder, where it stays as a record of what the gate missed.

\paragraph{The coverage index} The index reads each trial's metadata, groups trials into cells by the coverage axes, and assigns each trial one of the three classes of Equation~\eqref{eq:class}. The runner checks each trial's metadata against the contract when it writes the file and stores any problems there, without rejecting the trial, because the trial has already happened. A trial is \emph{unverified} when its record lacks a required field or holds a value that the contract does not declare: the data may be fine, but the record does not prove it. Most such trials were recorded before the contract existed, and the index shows them but never counts them. In one of our laboratory experiments, a cell with 14 trials, 11 of which are excluded, provides 3 covered trials, not 14. The index also lists level combinations with no trials, and for a campaign, the planner reports how many cells of Equation~\eqref{eq:cells} are wanted, covered, and still missing.

\paragraph{Computers, agents, and interfaces} A factor level can select the computer that runs the trial. The runner reaches it over SSH, records in its container, and copies a reduced recording without images back to the laptop. A trial that records nothing receives no metadata, so an empty directory never fills a cell. Each trial agent is explicit: each computer is a level of the host factor with its own gate checks, each robot is an identity field in the contract, and the metadata stores the operator's confirmations, overrides, and stops. This separates data by agent, as Ortega et al. ask for \cite{ortega2026replicable}. A named analysis selects its trials from the index and lists the exact recordings it used, so a figure can be traced back to its trials.

\paragraph{Requirements and scope} The framework needs three things from a system: a command that starts its software, including a simulator if there is one, a recorder that the runner starts and stops, and a script with an exit code for each condition to check. The planner, the linter, the gate, and the coverage index are in a general package that names no robot and no middleware, which tests enforce, with 8 general checks such as disk space. A second package connects it to ROS 2, where a recording is a bag, with 13 more checks such as topic rates, and optional adapters for the Crazyflie, Spot, and PX4 hold six checks and two verify steps. All four studies use ROS 2. Without ROS, one of our Crazyflie drones flew one hover through its Python library, with the drone's own log as the recording, and the general package ran the gate, the verify step, and the coverage index without any change. We have not ported a study to another middleware.

\section{Evaluation}
\label{sec:eval}

We evaluate the framework in four studies. Study~1, the laboratory study, asks whether the trials of three experiments with real robots became traceable and complete with the framework (Section~\ref{sec:rq4}). Studies~2 and~3 test the three problems of Section~\ref{sec:intro}: Study~2, the fault study, injects faults that the software does not report and checks a lighting level that the operator sets (Section~\ref{sec:rq1}), and Study~3, the team study, varies the number of robots, 1 to 3, and makes one agent among 5 to 20 faulty (Section~\ref{sec:rq2}). Study~4, the LLM study, tests whether someone else, here an LLM, can write the declarations and run them (Section~\ref{sec:rq3}). The last three run in simulation, and we repeated the fault study, the first team campaign, and the lighting check on real robots, where we report what the framework detected and recorded, not the robots' performance.

\subsection{Setup and Pre-Registration}
\label{sec:setup}

\paragraph{Robots and experiments} We collect hardware data in our laboratory, where a quadruped robot and nano-UAVs run joint localization experiments, using motion capture as ground truth. The robots are a Boston Dynamics Spot with an arm and Bitcraze Crazyflie drones, and the software runs on three computers: a laptop, a lab PC, and an embedded GPU computer on the quadruped. Three experiments with real robots use the framework. In the \emph{mocap experiment}, run from the laptop, the drone flies on the motion-capture position alone. In the \emph{static-camera experiment}, a static depth camera localizes the drone. In the \emph{arm-camera experiment}, the quadruped's arm camera localizes the drone \cite{loritemora2026quadruped}. The two camera experiments run on the quadruped's computer. An \emph{example experiment} without a robot, a ROS 2 talker that the package includes, runs on the laptop and the lab PC.

\paragraph{Simulated quadrotor} Except for the fleet campaign of Section~\ref{sec:rq2}, the simulated studies use a PX4 quadrotor in the Gazebo simulator, whose airframe, autopilot, and ROS 2 bridge differ from those of our laboratory drones. Its missions are a hover or a square at 2.0\,m, in still air or in wind.

\paragraph{Pre-registration} For the fault, team, and LLM studies, we committed the hypotheses, the predicted outcomes, and the analysis plan before the main runs, log every deviation, and release these files with the code.

\subsection{Study 1: Records in a Real Laboratory}
\label{sec:rq4}

\paragraph{Design} The population consists of all recordings from the three experiments with real robots on the laptop from March to September 2026, 261 after removing duplicates. The cut is the first commit in which the runner writes a trial's metadata at capture, with 174 recordings before it, the baseline, and 87 after it. A recording is \emph{traceable} in the sense of Ortega et al. \cite{ortega2022testing} when a machine-readable record names it, either its own metadata file or another file such as a manifest. It is \emph{complete} when it is usable under today's contract or excluded with a recorded reason. The index holds only recordings with a metadata file, so it contains fewer trials than there are recordings.

\begin{figure}[htbp]
    \centering
    \includegraphics[width=0.8\columnwidth]{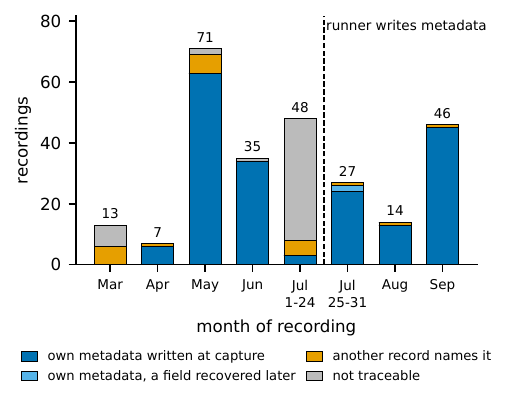}
    \Description{A stacked bar chart of the laboratory's recordings per month from March to September 2026, with July split into the 1st to 24th and the 25th to 31st. Each bar has four parts: own metadata written at capture, own metadata with a field recovered later, named by another record, and not traceable. The totals are 13, 7, 71, 35, 48, 27, 14 and 46. A dashed line between the two parts of July marks the first commit where the runner writes the metadata. Before it, many recordings are not traceable or only named by another record. After it, almost every recording has its own metadata written at capture.}
    \caption{Recordings of the laboratory per month, by how their metadata was written. The dashed line marks the first commit that wrote a trial's metadata at capture. Capture-time metadata becomes the rule after the cut.}
    \label{fig:rq4}
\end{figure}

\paragraph{Results} The large effect is completeness, which rose from 6.3\,\% before the cut to 86.2\,\% after it. Its low value before the cut is partly expected, because the contract did not exist yet, so it measures how many recordings can prove their validity today. Traceability rose from 71.3\,\% to 100\,\%, and the share of recordings whose own metadata was written at capture rose from 60.9\,\% to 94.3\,\% (Figure~\ref{fig:rq4}). The traceability gain is not robust. The laboratory notebooks already recorded metadata, and the low share before the cut comes mostly from the first month and from hand-recorded sessions. With the hardware scratch recordings included, the difference becomes small and not significant.

The verify steps also rejected 10 trials in which the action exited normally, while the bag showed that the observer node published nothing or ran below 5\,Hz. A retrospective run over older bags found two unmarked position-only flights whose drone stayed on the ground. At the time of the study, the index held 202 trials of the three experiments and the example experiment, and fewer than a third of them were usable.

Counting only usable trials also changes results computed from real data. With a named analysis, we scored the vision estimate of the drone in every recorded trial of the two camera experiments as the median 3D error against motion capture. In the one cell of the arm-camera experiment whose 15 trials share a detector and a pose source, counting every trial raises the mean error from 71 to 76\,mm (+7\,\%, 95\,\% bootstrap interval 4 to 9\,\%). Its two unverified trials ran a controller tuning that the contract did not declare, and its two excluded trials were interrupted in the air. In the static-camera experiment, which has only 3 usable trials with data, the comparison shows no clear difference. The trials of the arm-camera experiment in which the drone carried AprilTag markers \cite{loritemora2026quadruped} give a naive mean error, although none of them is usable, so the framework reports no value for them.

\subsection{Study 2: Injected Faults and the Checks}
\label{sec:rq1}

\paragraph{Design} The campaign crosses a \texttt{gate} factor, blocking or non-blocking, with a \texttt{fault} factor of seven levels: six faults that reproduce failures documented in our laboratory (a complete leftover stack of the other world, a leftover simulator world without an autopilot, a ground-truth bridge that dies, a middleware configuration that cannot discover the autopilot bridge, a vehicle that is not ready to arm, and an unwritable bag directory), and a control without a fault. A non-blocking gate runs every check and stores its verdict (Listing~\ref{lst:gate}) but never stops a trial, so the two arms differ only in the blocking decision. The gate under test is the one used in the experiment before this study, with checks for four of the six faults. The 80 attempts ran unattended in five randomized blocks, and an oracle that does not use the framework (a process witness, the autopilot's own log, and a separate bag reader) classified every attempt (Table~\ref{tab:rq1}).

\begin{table}[htbp]
    \caption{Outcome of the 80 attempts of the fault study, per fault and gate arm. O1 blocked at the gate, O2 failed without metadata, O3 aborted by the exit code, O4 aborted by a verify step, O5 counted and valid, O6 counted but silently wrong. NO-GO: the gate's verdict over both arms.}
    \label{tab:rq1}
    \footnotesize
    \setlength{\tabcolsep}{2.4pt}
    \begin{tabular}{@{}lcccccccccccccc@{}}
        \toprule
        & \multicolumn{6}{c}{blocking gate} & & \multicolumn{6}{c}{non-blocking gate} & \\
        \cmidrule(lr){2-7}\cmidrule(lr){9-14}
        Fault & O1 & O2 & O3 & O4 & O5 & O6 & & O1 & O2 & O3 & O4 & O5 & O6 & NO-GO \\
        \midrule
        none & 0 & 0 & 0 & 0 & 10 & 0 & & 0 & 0 & 0 & 0 & 10 & 0 & 0/20 \\
        leftover stack & 0 & 0 & 0 & 0 & 0 & 5 & & 0 & 0 & 0 & 0 & 0 & 5 & 0/10 \\
        stale world & 0 & 0 & 0 & 0 & 0 & 5 & & 0 & 0 & 0 & 0 & 0 & 5 & 0/10 \\
        bridge down & 5 & 0 & 0 & 0 & 0 & 0 & & 0 & 0 & 0 & 5 & 0 & 0 & 10/10 \\
        no discovery & 5 & 0 & 0 & 0 & 0 & 0 & & 0 & 0 & 5 & 0 & 0 & 0 & 10/10 \\
        not ready & 0 & 0 & 5 & 0 & 0 & 0 & & 0 & 0 & 5 & 0 & 0 & 0 & 0/10 \\
        bag dir. & 5 & 0 & 0 & 0 & 0 & 0 & & 0 & 5 & 0 & 0 & 0 & 0 & 10/10 \\
        \bottomrule
    \end{tabular}
\end{table}

\paragraph{Results} Every cell met its frozen prediction (Table~\ref{tab:rq1}), and every silent error occurred in a factor whose check was missing or not sound, as Equation~\eqref{eq:silent} implies. With the blocking gate, the three faults whose checks fired were blocked in all 15 attempts, and none produced a bag. The fourth checked fault, the leftover stack, passed the gate in 5 of 5 attempts because the gate counted one process of each kind, while the mission flew the leftover vehicle in the other world. The leftover world without an autopilot had no check and was silently wrong 10 of 10 times. Over the four checked faults, 5 of 20 attempts were silently wrong in each arm, so our hypothesis that the gate reduces silent errors is not supported in this design (one-sided Fisher exact test, $p=0.64$). With the non-blocking gate, a later layer stopped each of these three faults: the no-metadata rule, the exit code, or a verify step. For these faults, the gate's measured benefit is that no bag is written and that the attempt ends sooner. The gate never rejected a clean attempt (0 of 20) and took a median of 12.7\,s.

Silent errors also bias the results. The silently wrong hovers have larger tracking errors than the valid ones, so the hover cell, as the index counts it, reads 28\,\% above its valid trials. In an exploratory follow-up, knowing the faults, we added two missing checks: one that compares the running simulator world with the declared one, and one that waits for the autopilot's own preflight check. The world check added the simulator world to $F_G$, and the two checks blocked all 15 attempts of the three faults that had passed.

\paragraph{On hardware} We repeated four of the checked faults with the real drone of the mocap experiment, twice each, with the blocking gate. The gate refused all 8 attempts before recording anything, each on the check written for its fault. During three of these attempts, motion capture also lost the drone, a fault that nobody injected, and the gate recorded both reasons. In a campaign with the two harmless faults and a non-blocking gate, the drone flew 4 times, and 2 trials were verified and counted. On the same day, the checks also stopped 4 faults that nobody had planned: a drone motor that did not start (a verify step measured almost no vertical travel), a battery just below its threshold, drones that were switched off, and a stalled camera stream. Each of them is a line in the gate log with its measured value.

\paragraph{Closing the gaps} The hardware day also exposed two gaps in our own checks (Section~\ref{sec:discussion}), closed six days later: the gate now reads the operator's stop latch with the same 120\,s rule as the trial script, and the team action confirms from the pose stream that each drone reached its takeoff altitude before the next takeoff, or it lands every drone and aborts. In a pre-registered simulated campaign of 15 attempts, the gate blocked all 5 attempts with a fresh latch, the action aborted all 5 with a grounded lead (a dropped takeoff command, since the simulator has no motor model), and all 5 controls counted.

\paragraph{Lighting} Lighting is easy to vary in a simulator and hard to vary in a laboratory, so we treat it as a factor in both domains, a Gazebo world in simulation and an operator step on hardware. In both, a gate check compares the camera image's mean gray level (0 to 255) with a band for the declared level, calibrated and committed before the first trial, and the linter refuses overlapping bands. In simulation, a quadrotor hovered twice at each of four light levels with disjoint bands, all 8 trials passed their check, and the gate refused a deliberate mismatch (dark declared, bright world running).
On hardware, with the arm camera of the arm-camera experiment, the camera's automatic exposure made the lights-on and lights-off levels measure almost the same, so the linter would have refused the campaign. With a fixed exposure, the two levels were clearly apart, and the gate refused the deliberate mismatch. The first attempt that would have counted was refused because the camera stream stalled, and we ended the block there. No hardware trial was counted, so every level stays in $H_{\text{lighting}}$ of Equation~\eqref{eq:hgap}, as the index reports.

\subsection{Study 3: The Number of Robots as a Factor}
\label{sec:rq2}

\paragraph{Design} Two campaigns treat the number of robots as a factor. The first adds \texttt{n\_vehicles}, 1, 2, or 3 simulated PX4 quadrotors in one world, crossed with the mission (hover or square), in four randomized blocks, 24 trials and 48 flights, and tests whether each recorded agent's onboard estimate is closest to the body that the metadata names. The fleet campaign also runs in Gazebo, with $N = 5$, 10, or 20 robots, all nano-UAVs with a model of our laboratory's drone or three of them per ground robot, and makes one agent, drawn at random, faulty in one of four ways that we also observed with real robots (stale pose, wrong mission, lost bridge, never moves) or none. Each agent is its own process, and a fleet watchdog, the action's start check, and two verify steps compare the agents with the declaration. We predicted the layer that refuses each fault and that its record names the faulty agent, and ran 180 attempts with the blocking gate in five randomized blocks.

\paragraph{Results in simulation} In the first campaign, all 24 trials are usable, all 48 flights were attributed to the correct vehicle, and the framework added less than 1\,s per vehicle. In the fleet campaign, 179 of 180 attempts met their frozen prediction (Table~\ref{tab:rq2}): all 60 controls counted, all 120 faulty attempts were refused, 119 on the declared layer, and the record named exactly the faulty agent in 115 of 120 refusals and contained its name in all 120. The other five named a second agent, three of them after a ground robot on its wrong mission pushed a neighbor. Neither refusal nor exact naming depended clearly on $N$ or on the fleet composition. The gate's duration and the trial metadata grew only slightly with each agent. The one failed prediction is the simulator's real-time factor at $N = 20$, below the predicted 0.8 in 10 of 50 attempts on a laptop shared with another campaign.

\begin{table}[htbp]
    \caption{The fleet campaign per fault, pooled over $N$ and fleet composition, with the outcomes of Table~\ref{tab:rq1} and the exact 95\,\% interval of the attribution.}
    \label{tab:rq2}
    \footnotesize
    \setlength{\tabcolsep}{2.4pt}
    \begin{tabular}{@{}lccccl@{}}
        \toprule
        Fault & Attempts & Predicted & As predicted & Refused & Named exactly \\
        \midrule
        none & 60 & O5 & 60 & 0 & n/a \\
        stale pose & 30 & O4 & 29 & 30 & 29/30 (0.83 to 1.00) \\
        wrong mission & 30 & O4 & 30 & 30 & 27/30 (0.73 to 0.98) \\
        dead bridge & 30 & O1 & 30 & 30 & 29/30 (0.83 to 1.00) \\
        never moves & 30 & O3 & 30 & 30 & 30/30 (0.88 to 1.00) \\
        \bottomrule
    \end{tabular}
\end{table}

\paragraph{On hardware} With real drones on motion capture, the single-drone level ran its 3 trials and the two-drone level ran 2 of its 3, all verified, with each drone attributed to its own body within 3\,mm. The remaining cells were not flown, and the planner lists them as missing. Of the 17 attempts, 5 were counted, and the ledger holds the reasons for the other 12.

\subsection{Study 4: Experiments Written by LLMs}
\label{sec:rq3}

This study has two parts: a controlled comparison of the designs that LLMs write, and an end-to-end adoption run.

\paragraph{Design comparison} The first part gives an LLM ten experiment requests in plain language, six for the simulated quadrotor and four for the example experiment, each with a correct design written before any model call. Two are infeasible on purpose. In the declaration arm, the model writes the declarations and receives output from four validators (the linter, the campaign checker, the planner, and a dry run) for up to three rounds, and its first round is scored as the no-feedback condition. In the script arm, the model writes a free-form script that launches the stack, records, and logs the conditions, with documentation of nearly the same length. Two Claude models, \texttt{claude-sonnet-5} and \texttt{claude-haiku-4-5-20251001}, answered every task twice in each arm, 80 sessions in total. An automatic scorer, not an LLM, checks whether each answer runs, whether its factors, levels, and repetitions match the request exactly, whether every trial's condition can be recovered from the recording, and whether it contains a silent error. One replicate of every runnable answer then ran in simulation.

\paragraph{Results of the comparison} The declarations matched the request exactly in 32 of 32 answers for feasible tasks and the free-form scripts in 27 of 32 (Table~\ref{tab:rq3}), a difference unlikely to be chance (exact McNemar test, one-sided as pre-registered, $p=0.031$). Still, four of the five differences come from one task, the request for every combination of world, mission, and altitude, so the result rests mainly on that task. The one silent error is a script that flew at 4.0\,m instead of 1.0\,m and reported nothing. In a declaration, this error cannot go unnoticed: a new level must appear in the factor file, the campaign checker refuses an undeclared level, and Equation~\eqref{eq:lint} forces every swept factor into the record. Every declaration passed the validators on its first round, so these tasks do not support the hypothesis that feedback improves designs. On the two infeasible tasks, 6 of 8 declarations and 3 of 8 scripts refused the request, and no answer realized an impossible level. We do not read it as better programming: the factor file fixes the levels, the runner writes the record, and the validators refuse an incomplete experiment.

\begin{table}[htbp]
    \caption{Static scores of the design comparison per model and condition. Runnable and exact scores are computed over the 16 answers to feasible tasks, and silent errors over all 20 answers. The declarations row stands for both the first round and the feedback condition, which are identical because no declaration needed a second round.}
    \label{tab:rq3}
    \footnotesize
    \setlength{\tabcolsep}{4pt}
    \begin{tabular}{@{}llccc@{}}
        \toprule
        Model & Condition & Runnable & Exact design & Silent error \\
        \midrule
        Sonnet & declarations & 16/16 & 16/16 & 0/20 \\
        Sonnet & free-form script & 16/16 & 14/16 & 1/20 \\
        Haiku & declarations & 16/16 & 16/16 & 0/20 \\
        Haiku & free-form script & 13/16 & 13/16 & 0/20 \\
        \bottomrule
    \end{tabular}
\end{table}

Execution in simulation gave no fair comparison: all but one of the declared designs for the simulated quadrotor ran, with every condition recoverable from the recording, while every script failed at its first command on strict shell options that the ROS 2 Humble setup file rejects, a reason unrelated to the design. The declared designs for the example experiment did not run, because a defect in that experiment's own gate refused them, which we fixed afterward.

\paragraph{Adoption from the documentation} We could not run a usability study with other people, so the second part is an automated adoption run as a proxy. An LLM coding agent (Claude Opus 5.5, \texttt{claude-opus-5-5}) received only the user guide, the conventions file, and the package and launch file of a simulated legged robot (Spot with the CHAMP controller in Gazebo), and no access to the history of our experiments. Its task was to declare an experiment in which the robot walks a 1\,m square at two speeds with two repetitions each, with a verify step on the recording, and to run the campaign to full coverage. We committed the task and the measures before each run.

\paragraph{Results of the adoption run} A first agent reached full coverage (4 of 4 verified trials) after 35 minutes and reported six framework defects that we confirmed in the code, for example a gate check that refused the first trial of every new experiment. We fixed four of them, each with a test, and added two commands: one copies the nearest existing experiment, and one validates an experiment in seconds. A second fresh agent with the same task then reached full coverage after 26 minutes, without human help or refusal. Every gate and verify step passed. One automated subject per run is not a usability study, but it shows that an LLM agent can declare, from the documentation alone, a legged robot whose simulator and controller our laboratory does not use.

\section{Discussion and Limitations}
\label{sec:discussion}

The first three studies answer the three problems of Section~\ref{sec:intro}. For the operator, the gate stores every confirmation and override, and the lighting check gives an operator step a machine check, although no hardware trial with it counted yet. For several agents, the record that the contract requires attributed every simulated flight to the correct vehicle and named exactly the faulty agent in almost every refusal. For a process that ends without error, the verify steps rejected laboratory trials whose software had exited normally and caught a drone motor that did not start, and the coverage index counts only usable trials: counting every trial would have raised the error of one condition by 7\,\%.

A gate checks exactly what it declares, and its log makes every refusal countable. In the fault study, blocking at the gate did not reduce silent errors, because a later layer stopped every fault that the gate's checks caught. The two faults that passed, one whose check did not fire and one without a check, passed every later layer as well, and only the gate checks that we added afterward stopped them, as Equation~\eqref{eq:silent} implies. The gate therefore matters for correctness where no later layer sees a wrong condition, such as the simulator world or the lighting, and elsewhere it saves time. We wrote the added checks knowing the faults, so this shows that a failure can become a check, not how well a gate detects unknown failures. This is the loop the framework supports: every observed failure becomes a check.

In the fault, team, and lighting studies the same declaration ran first in simulation and then on hardware, and the index reports which levels have evidence only from simulation (Equation~\eqref{eq:hgap}). This is the practice we recommend: simulation first, and hardware for what simulation cannot answer.

The work has four limitations. First, we built the framework and also designed its tests, including the faults and the LLM tasks. We tied each fault to a documented failure and committed the predictions before the main runs, but changed two of them after a pilot. Second, only we have used it: one operator at one site, with an LLM agent as a stand-in for a new user, so its usability for other people is unknown. Third, the studies are small and narrow: one laboratory, one flight software in simulation, ten LLM tasks, and only ROS 2, with a single drone hover without ROS and no use outside robotics. The laboratory study also compares an early period with a later one, when our practice had matured as well, so not all of the gain is due to the framework. Fourth, the records are flat files per trial rather than a provenance graph, and the gate records that the operator confirmed an item but not what the operator observed.

\section{Conclusion}
\label{sec:conclusion}

Experiments with robot teams need a record of each trial's condition and validity, and a record kept by hand is easily lost. RoboFACTS lets a researcher declare an experiment and leaves planning, pre- and post-trial checks, and counting usable trials to the software, in simulation and on hardware. In our laboratory, its checks rejected trials whose software had exited normally and refused every faulty attempt on real robots, and the share of recordings with a complete record rose from 6.3\,\% to 86.2\,\%. The lesson for other laboratories is to declare what each trial must record, validate the declaration in simulation, check it at the trial, count only the trials that pass, and turn every observed failure into a new check. The framework is open source, and in future work we will export the records to PROV-O and study how a second laboratory uses it.

\begin{acks}
This research was supported by the IT University of Copenhagen and Helix Lab in Denmark under the Novo Nordisk Fonden through project grant number 224611. Generative AI use: LLM coding agents drafted parts of this text, the analysis scripts, and parts of the framework's code from the authors' specifications, and the authors reviewed, rewrote, and tested them.
\end{acks}

\bibliographystyle{ACM-Reference-Format}
\bibliography{references}
\end{document}